\documentclass[conference]{IEEEtran}
\IEEEoverridecommandlockouts

\usepackage{cite}
\usepackage{amsmath,amssymb,amsfonts}
\usepackage{algorithmic}
\usepackage{graphicx}
\usepackage{textcomp}
\usepackage{xcolor}
\usepackage{acro}
\usepackage{float}
\usepackage{footnote}
\usepackage{tabularx}
\usepackage{pifont}
\usepackage{booktabs}
\usepackage{threeparttable}
\usepackage{multirow}
\usepackage{xurl}
\usepackage{subfigure}
\usepackage[colorlinks=true, allcolors=black]{hyperref}

\newcommand{\xmark}{\ding{55}}

\def\BibTeX{{\rm B\kern-.05em{\sc i\kern-.025em b}\kern-.08em
    T\kern-.1667em\lower.7ex\hbox{E}\kern-.125emX}}

\DeclareAcronym{ebike}{
    short = E-Bike,
    long  = Electric Bike,
}
\DeclareAcronym{gdpr}{
    short = GDPR,
    long = General Data Protection Regulation,
}
\DeclareAcronym{ecar}{
    short = E-Car,
    long  = Electric Car,
}
\DeclareAcronym{em}{
    short = EM,
    long  = Electric Micromobility
}
\DeclareAcronym{escooter}{
    short = E-Scooter,
    long  = Electric Scooter,
}
\DeclareAcronym{ev}{
    short = EV,
    long  = Electric Vehicle,
}
\DeclareAcronym{gps}{
    short = GPS,
    long  = Global Positioning System
}
\DeclareAcronym{mae}{
    short = MAE,
    long  = Mean Absolute Error,
}
\DeclareAcronym{ml}{
    short = ML,
    long  = Machine Learning,
}
\DeclareAcronym{roi}{
    short = RoI,
    long  = Region of Interest
}
\DeclareAcronym{sem}{
    short = SEM,
    long  = Shared Electromobility
}
\DeclareAcronym{soc}{
    short = SoC,
    long  = State of Charge,
}
\DeclareAcronym{ai}{
    short = AI,
    long  = Artificial Intelligence,
}
\DeclareAcronym{lr}{
    short = LR,
    long  = Linear Regression,
}
\DeclareAcronym{svr}{
    short = SVR,
    long  = Support Vector Regression,
}
\DeclareAcronym{dt}{
    short = DT,
    long  = Decision Tree,
}
\DeclareAcronym{rf}{
    short = RF,
    long  = Random Forest,
}
\DeclareAcronym{gb}{
    short = GB,
    long  = Gradient Boosting,
}
\DeclareAcronym{knn}{
    short = KNN,
    long  = K Nearest Neighbour,
}
\DeclareAcronym{xgb}{
    short = XGB,
    long  = eXtreme Gradient Boosting,
}
\DeclareAcronym{lgbm}{
    short = LGBM,
    long  = Light Gradient-Boosting Machine,
}
\DeclareAcronym{mlp}{
    short = MLP,
    long  = Multi-Layer Perceptual,
}

\begin{document}

\title{Data-driven Energy Consumption Modelling for Electric Micromobility using an Open Dataset\\

\author{Yue Ding, Sen Yan, Maqsood Hussain Shah, Hongyuan Fang, Ji Li and Mingming Liu \thanks{Y. Ding and H. Fang are with the SFI Centre for Research Training in Machine Learning (ML-Labs) at Dublin City University. S. Yan, M. H. Shah and M. Liu are with the SFI Insight Centre for Data Analytics and the School of Electronic Engineering, Dublin City University, Dublin, Ireland. J. Li is with the Department of Mechanical Engineering, University of Birmingham.

Y. Ding and S. Yan are joint first authors. \textit{Corresponding author: Mingming Liu. Email: {\tt mingming.liu@dcu.ie}.}}}}

\maketitle

\begin{abstract}
The escalating challenges of traffic congestion and environmental degradation underscore the critical importance of embracing E-Mobility solutions in urban spaces. In particular, micro E-Mobility tools such as E-scooters and E-bikes, play a pivotal role in this transition, offering sustainable alternatives for urban commuters. However, the energy consumption patterns for these tools are a critical aspect that impacts their effectiveness in real-world scenarios and is essential for trip planning and boosting user confidence in using these. To this effect, recent studies have utilised physical models customised for specific mobility tools and conditions, but these models struggle with generalization and effectiveness in real-world scenarios due to a notable absence of open datasets for thorough model evaluation and verification. To fill this gap, our work presents an open dataset, collected in Dublin, Ireland, specifically designed for energy modelling research related to E-Scooters and E-Bikes. Furthermore, we provide a comprehensive analysis of energy consumption modelling based on the dataset using a set of representative machine learning algorithms and compare their performance against the contemporary mathematical models as a baseline. Our results demonstrate a notable advantage for data-driven models in comparison to the corresponding mathematical models for estimating energy consumption. Specifically, data-driven models outperform physical models in accuracy by up to 83.83\% for E-Bikes and 82.16\% for E-Scooters based on an in-depth analysis of the dataset under certain assumptions.
\end{abstract}


\section{Introduction}
Nowadays, micromobility is attracting significant attention worldwide for its potential to help establish a sustainable, cost-effective, and widely accessible modern transportation network \cite{d2023sustainable}. \ac{em}, including e-scooters and e-bikes, is set to be a key contributor to effectively mitigate first and last-mile problems by connecting various modes of the transportation network \cite{van2022preferences}. 

However, range anxiety remains a significant concern pertaining to the utilization of any e-mobility tool, undermining user confidence and hindering the widespread adoption of e-mobility as a preferred mode of transportation \cite{rangeanxietypevec2020survey}. To mitigate this concern, it is paramount to gain a comprehensive understanding of energy consumption patterns and, consequently, establish reliable methods for estimating energy consumption. This is crucial in facilitating trip planning for users, thereby enhancing their satisfaction with the services, particularly within the context of shared mobility. In this regard, some recent studies have used mathematical modelling techniques. However, these approaches are often challenging to prove their effectiveness in real-world scenarios due to the lack of an open dataset for further model evaluation and verification \cite{miri2021electric,xie2020microsimulation}. Recent advancements in AI have positioned ``data" at the forefront of progress across various vertical domains. By harnessing relevant data effectively, we can undertake diverse prediction tasks. For instance, data-driven approaches have become pivotal in energy modelling tasks, however, in the context of e-mobility in general and micro e-mobility in particular, there is an apparent gap which calls for open datasets for energy modelling related studies \cite{Muli2022, Yan2023}.

As per \cite{Riderguide2023}, the manufacturer-reported maximum travel ranges exceed owner-reported averages by about 30\%. This discrepancy stems from inadequate energy prediction practices within these companies, leading to ineffective monitoring and control of vehicle battery usage and energy efficiency. Consequently, it creates a disparity between advertised and actual performance, undermining consumer trust and satisfaction as experiences fail to meet expectations. However, current energy models \cite{Burani2022}, \cite{Wang2021} simplify the complex energy dynamics of these systems, often missing crucial details. Moreover, studies \cite{Wang2021, Yuniarto2022, Genikomsakis2017} in energy modelling lack detailed insights into travel patterns and data-driven methods which are essential to improve the accuracy and effectiveness of energy consumption predictions. We note that only a limited number of studies have primarily focused on comprehensive datasets pertaining to micromobility. The work \cite{Schu2023}  indicates a gap in current research dedicated to dataset development and refinement in this specific domain. 
Even when available, current datasets \cite{Chicago2019, Chicago2020, Wang2021} often lack details such as rider weight, terrain type, State of Charge (SoC), and weather conditions, which can affect the efficacy of energy modelling for various e-mobility tools. 
Moreover, the specifications provided by electric micromobility manufacturers are often not precise, particularly when it comes to their products' maximum travel range, as this is typically assessed under optimal and ideal conditions, making it challenging for users to estimate the energy needed for real-world journeys.   

 These aforementioned limitations in current work urgently underscore the need for comprehensive and openly accessible datasets pertaining to \ac{escooter}s and \ac{ebike}s, including crucial factors affecting energy consumption such as rider weight, terrain variations, and weather conditions. Furthermore, exploring a diverse range of modelling approaches beyond regression-based methods \cite{Wang2021}, \cite{en16031291} would contribute to a better understanding of energy consumption in the context of e-mobility frameworks. To bridge the identified research gaps, our work presents an openly accessible dataset covering both \ac{escooter}s and \ac{ebike}s, with meticulous consideration for critical factors in energy consumption. Our analysis extends beyond conventional regression-based models, incorporating diverse modelling techniques, including both \ac{ml} and deep learning, as well as its comparison with contemporary physical models. 


The rest of the paper is organised as follows. In \autoref{rw}, we provide a literature review of current datasets and related methods. In \autoref{emd}, we introduce the data collection and processing pipeline and provide an overview of the \ac{ebike} and \ac{escooter} datasets. In \autoref{ecm}, we present the models and describe the experimental results and analysis. The experiment results and relevant discussion are illustrated in \autoref{rd}. Finally, we sum up the discussions and limitations with a future pathway in \autoref{dl} and \autoref{cf}.

\section{Related Works} \label{rw}

In this section, we review multiple datasets pertaining to \ac{ebike} and \ac{escooter} in detail. We identify and discuss some limitations of these datasets and highlight how our work compares and contributes to the existing body of knowledge. A summary of the comparison is presented in Table \ref{tab:comparisonOfDatasets}.

\begin{table*}[htbp]
\label{tab:comparisonOfDatasets}
  \centering
  \caption{Comparison of Datasets in E-Mobility Research}
  \begin{tabularx}{\textwidth}{Xcccc}
    \toprule
    \textbf{Dataset} & \textbf{Coverage} & \textbf{Quality} & \textbf{Opensource} & \textbf{Focus} \\
    \midrule
    Chicago Pilot Programme \cite{Chicago2019} & \ac{escooter} trips, Chicago & Missing data, errors & \checkmark & Trip details, lacks energy metrics \\
    Micromobility Dataset \cite{Norfolk2020} & E-scooters and E-bikes & Lacks energy data & \checkmark & Trip details \\
    Gothenburg Dataset \cite{Wang2021} & \ac{escooter}, Gothenburg & Trajectories, energy data & \xmark & Energy consumption \\
    Moby Bikes Dataset \cite{moby-bikes} & E-bikes, Dublin & Errors, lacks energy data & \checkmark & Trip details \\
    \textbf{Our work} & E-scooters and E-bikes, Dublin & Trajectories, energy data & \checkmark & Trip details, energy consumption \\
    \bottomrule
  \end{tabularx}
\end{table*}

The existing datasets for \ac{em} studies are limited in both quantity and quality. One notable dataset \cite{Chicago2019} originates from the Chicago Pilot Programme, documenting \ac{escooter} trips capturing the commencement and conclusion of each journey during the 2019 initiative. This dataset was enhanced in 2020 with additional data columns pertaining to the vendor names. Despite its breadth, the dataset exhibits certain shortcomings including instances of missing data, non-descriptive trip identifiers, and errors such as identical start and end times. In effect, this dataset resembles an unprocessed data file, lacking in-depth trip details and omitting crucial energy consumption metrics. The Moby Bikes Dataset \cite{moby-bikes}, centred on \ac{ebike}s in Dublin, offers insights into trip details but is marred by \ac{gps} data error issues and the absence of energy consumption information. Similarly, the Micromobility dataset \cite{Norfolk2020}, encompassing \ac{escooter}s and \ac{ebike}s from July 2019 onwards, is updated weekly but echoes the shortcomings of the Chicago dataset with a notable absence of energy consumption or \ac{soc} information. 

Regarding energy consumption modelling, the recent works \cite{Yuniarto2022, Genikomsakis2017} primarily revolve around mathematical models that account for factors such as the braking system, energy load, etc. However, these models overlook the travel patterns of users. Moreover, there seems to be a gap in the integration of data-driven methods into these studies. This oversight could potentially limit the accuracy and applicability of the models, as user travel patterns can significantly influence energy consumption. The study in \cite{Wang2021} is relevant to our focus on energy consumption modelling which utilises Swedish \ac{escooter} data on timestamps, locations, and \ac{soc}. However, its dataset is not publicly available, which constrains broader research utility. The study's data collection lacks direct \ac{escooter} trajectory information, instead estimating travel metrics from origin-destination points with a minor time error due to data update frequency. Unlike our research that includes both \ac{escooter}s and \ac{ebike}s, it only considers \ac{ebike} data and also lacks a thorough comparison for different data-driven methods. It also overlooks key energy consumption factors like riders' physical attributes and weather conditions. 

\begin{figure*}[htbp]
    \vspace{-0.1in}
    \centering
    \includegraphics[width=\linewidth]{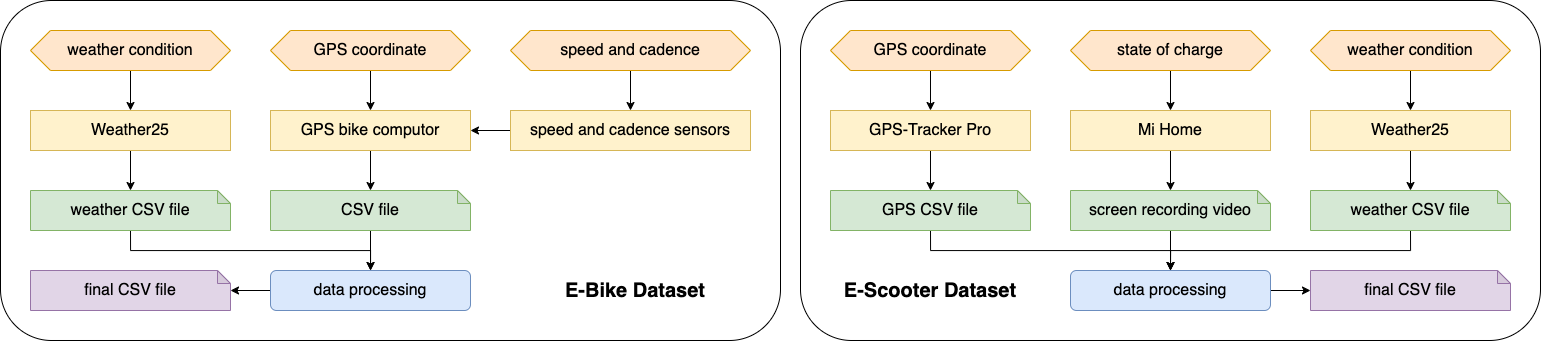}
    \vspace{-0.3in}
    \caption{The process of data collection and dataset creation.}
    \label{fig: datasets}
    \vspace{-0.1in}
\end{figure*}

Comparing with relevant works, a summary of our major contributions is outlined in the following:
    
    \begin{itemize}
        \item Performed thorough real-world data collection for both \ac{escooter}s and \ac{ebike}s, using a detailed and transparent methodology, resulting in the creation of an accessible \ac{gdpr}-compliant open dataset.         
        \item Performed an extensive analysis using various \ac{ai} techniques to model energy consumption and compared them with traditional physical models.
    \end{itemize}

\section{Electric Micromobility Dataset} \label{emd}

We introduce the data collection procedure and present our data processing flow for the dataset. The dataset will be used for further analysis in \autoref{ecm}.

\subsection{Pipeline of the E-mobility Trip Data Collection}
This section details the data collection process and encapsulates the process for the datasets. The complete process of data collection is wrapped up as a standard pipeline presented in \autoref{fig: datasets} and introduced in detail in the following sections.

\subsubsection{\ac{ebike}}

The data collection pipeline is listed below.

\noindent \textbf{Experiment Object:} For \ac{ebike}, we experimented with the model \textit{Electric Trekking Bike T1}\footnote{\url{https://eleglide.com/products/removable-battery-100km-range-electric-trekking-touring-bike-t1}}, a mountain bike equipped with a 250 W motor and capable of speeds up to 25 km/h. Powered by a 450 Wh battery, the \ac{ebike} has a range of 100 km and features a power assist module that offers five levels of assist/electric mode. These modes correspond to speeds of 12, 16, 20, 23, and 25 km/h. The dataset includes trip attributes for various pedal assist levels to mirror real-world scenarios.
        
\begin{figure}[htbp]
    \vspace{-0.1in}
    \centering
    \includegraphics[width=\linewidth]{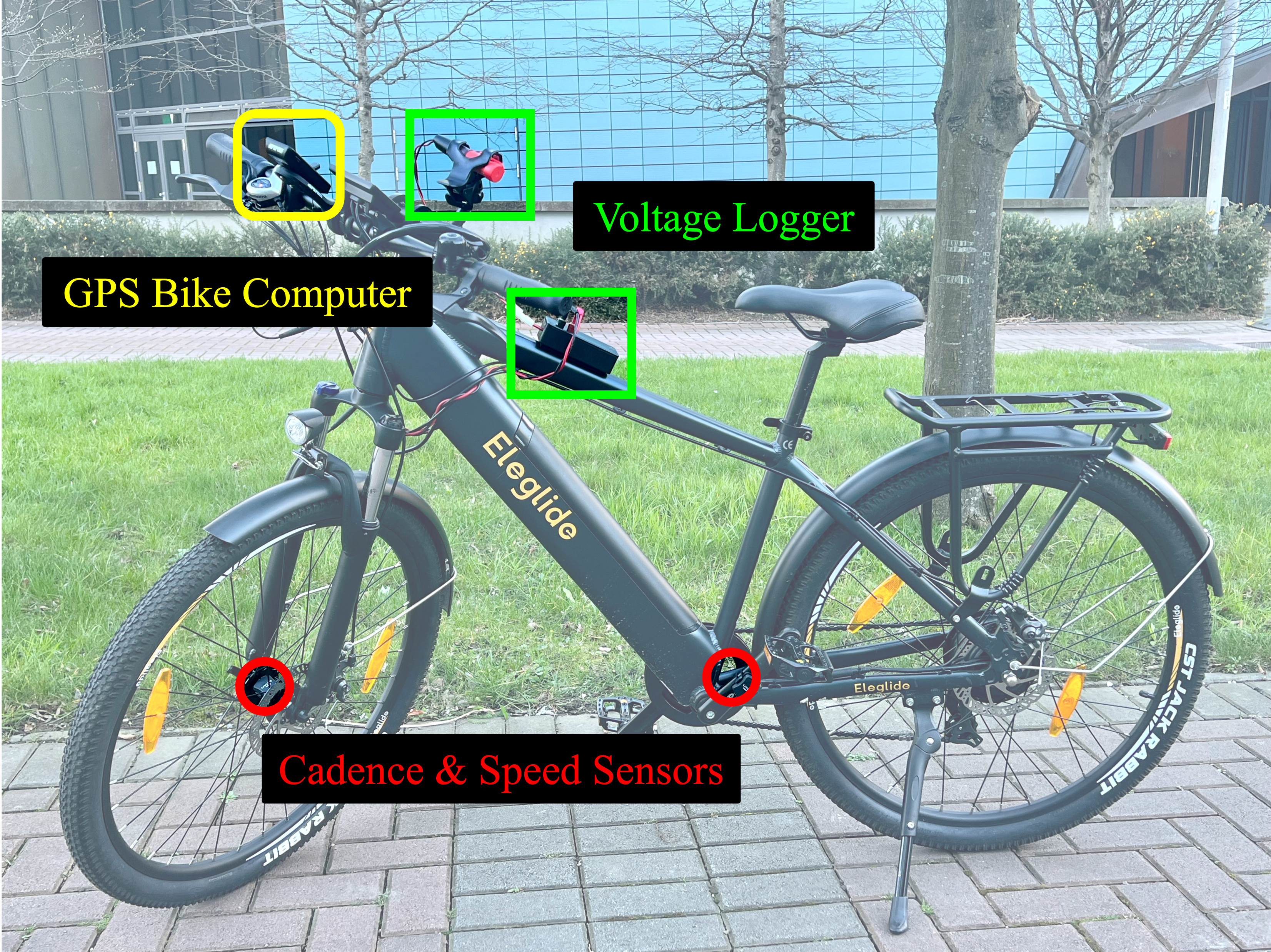}
    \vspace{-0.3in}
    \caption{\ac{ebike} and sensors for experimentation.}
    \label{fig: ebike}
    \vspace{-0.1in}
\end{figure}

\noindent \textbf{Data Collecting Device:} We employed an \textit{iGPSPORT iGS630 \ac{gps} bike computer}\footnote{\url{https://www.igpsport.com/igs630-highlights}} with \textit{LivLov V2 Bike Cadence and Speed Sensors}\footnote{\url{https://amzn.eu/d/j2OiRMi}} and the smart cycling application of the \ac{ebike} to gather journey attributes such as timestamps, \ac{gps} coordinates, altitude, speed, assistance level, and travel distance. In terms of \ac{soc} data, unfortunately, the \ac{ebike}'s built-in \ac{soc} display provides insufficient granularity (only 5 levels). This limitation is common in contemporary \ac{ebike}s. To address this, we employed a custom \textit{Voltage Logger} sensor\footnote{\url{https://ie.rs-online.com/web/p/data-loggers/1799537}} to continuously monitor real-time battery voltage. This sensor consists of a voltage-divider circuit, an \textit{ESP8266} microcontroller, and a relay for isolation, as shown in the upper green box in \autoref{fig: ebike}.  The microcontroller applies the voltage divider equation to calculate battery voltage and stores the data in a CSV file.

\noindent \textbf{Data Collection Process:} Before each \ac{ebike} trip, we activated the \ac{gps} bike computer, initialised the voltage logger, set the assistance level to a constant value, fixed the gear ratio, and commenced the ride. We stopped data collection at the trip's conclusion and switched off all the devices.

\subsubsection{\ac{escooter}}

The data collection pipeline is listed below.

\begin{figure}[htbp]
    \vspace{-0.1in}
    \centering
    \subfigure[Mi Home interface.]{
        \includegraphics[width=0.22\textwidth]{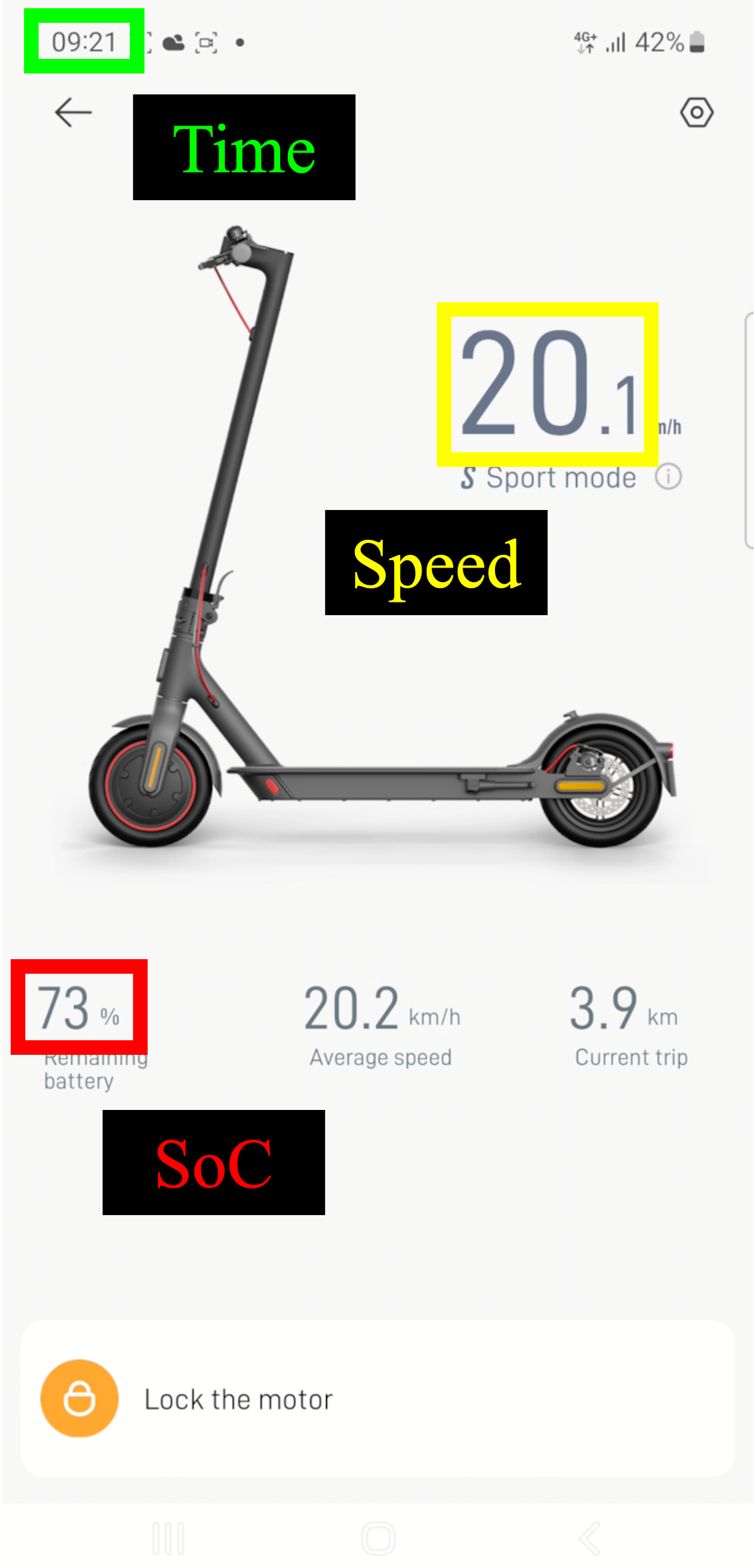}
        \label{subfig: mi home interface}
    }
    \subfigure[GPS-Tracker Pro interface.]{
        \includegraphics[width=0.22\textwidth]{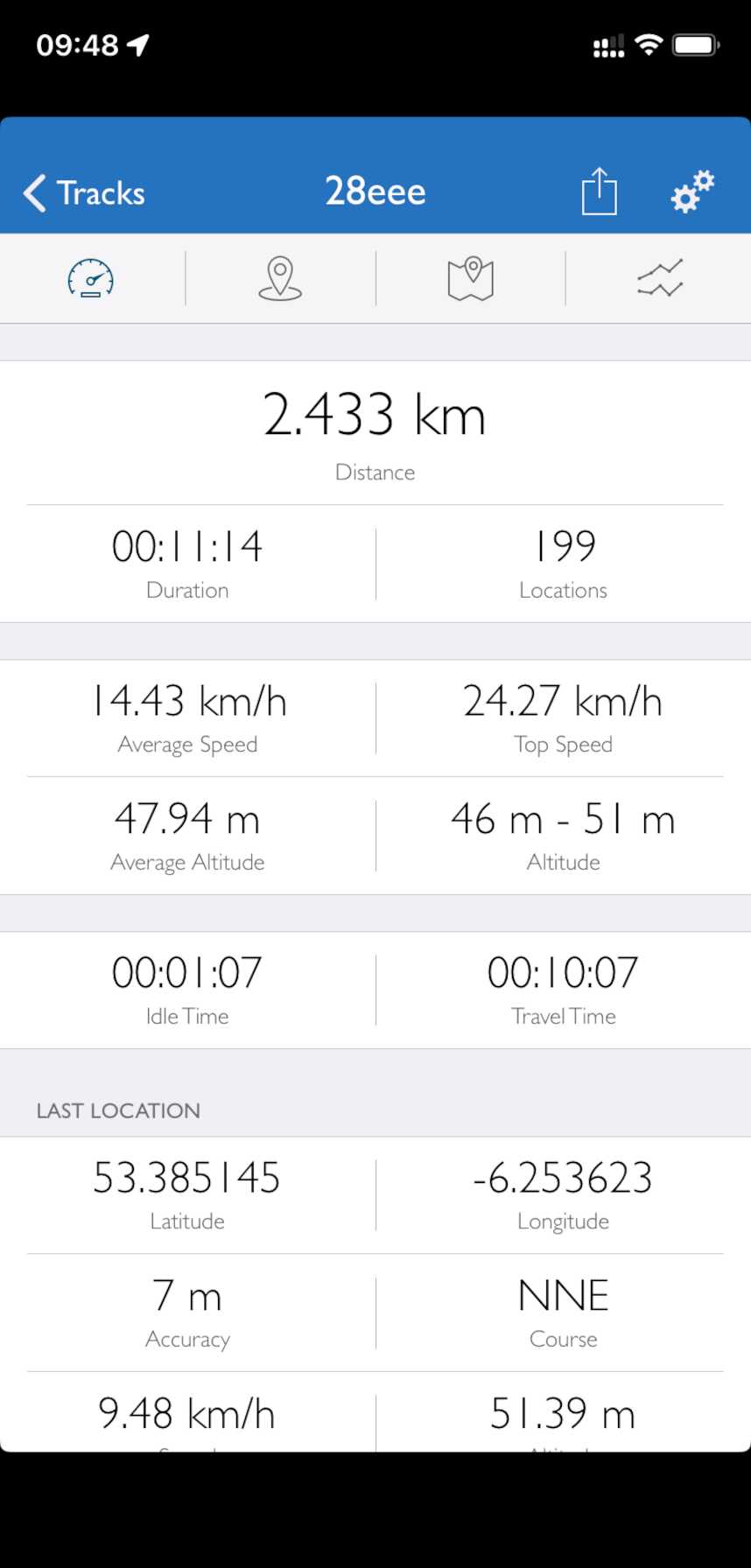}
        \label{subfig: GPS tracker interface}
    }
    \caption{Sample user interfaces of data collecting devices.}
    \label{fig: sample UI}
    \vspace{-0.1in}
\end{figure}

\noindent \textbf{Experiment Object:} For \ac{escooter} studies, we chose the popular \textit{Mi Electric Scooter Pro 2} model \footnote{\url{https://www.xiaomi.ie/mi-electric-scooter-pro-2}} due to its widespread availability and representative features. This model is equipped with a powerful 600-watt motor and offers three speed modes, each with its own speed range. During our data collection, we selected the sports mode, limiting the speed range to 0-25 km/h. The scooter is outfitted with a high-capacity 446 Wh lithium battery, providing a maximum range of 45 km on a full charge and a top speed of 25 km/h.
    
\noindent \textbf{Device Specification:} For the data collection device, two mobile applications were employed to gather attributes of \ac{escooter} trips, containing timestamp, \ac{gps} coordinates, altitude, speed, and \ac{soc}. The \ac{soc} was extracted from screen recordings using \textit{Xiaomi}'s official mobile application, \textit{Mi Home}\footnote{\url{https://play.google.com/store/apps/details?id=com.xiaomi.smarthome}}, as shown in the red box in \autoref{subfig: mi home interface}, on an Android device (\textit{SAMSUNG Galaxy A53}), while other attributes were directly acquired via a \ac{gps} tracking mobile application, \textit{\ac{gps}-Tracker Pro}\footnote{\url{https://apps.apple.com/us/app/gps-tracker-pro/id984920064}}, as shown in \autoref{subfig: GPS tracker interface}, on an Apple device (\textit{iPhone 11}). 

\noindent \textbf{Data Collection Process:} Before commencing each \ac{escooter} trip, we initiated the data collection applications on two mobile devices and concluded data collection upon trip completion. After multiple iterations of data collection, we proceeded to process and conduct a thorough analysis of the data gathered.

\subsection{Data Processing}

\subsubsection{\ac{ebike} Data}

The key steps for data processing are:

\noindent \textbf{Data Integration:} Data from both sources (the \ac{gps} bike computer and sensors) is merged into a unified file.

\begin{figure}[htbp]
    \vspace{-0.1in}
    \centering
    \includegraphics[width=\linewidth]{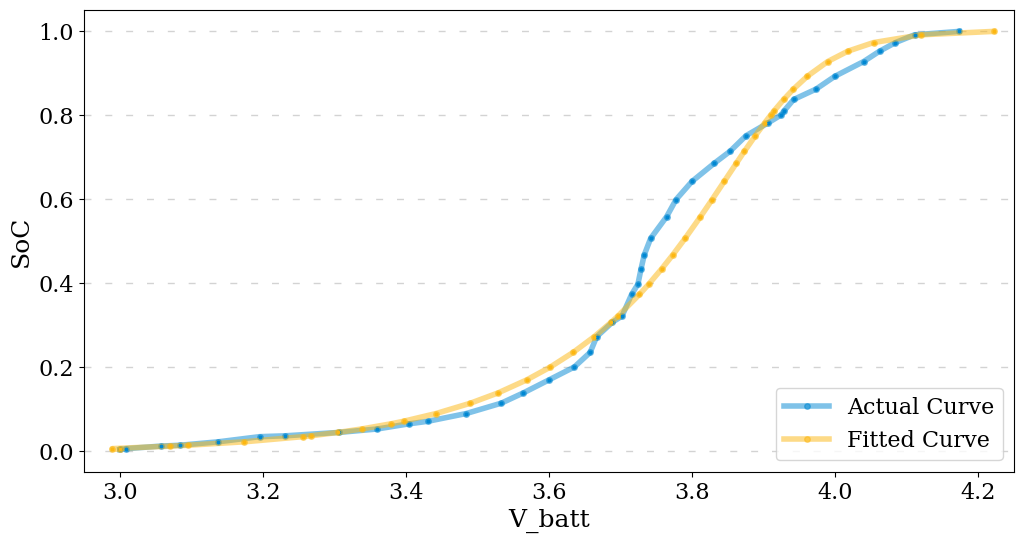}
    \vspace{-0.3in}
    \caption{Lithium polymer battery characteristics (SoC against voltage).}
    \label{fig: soc}
\end{figure}

\noindent \textbf{\ac{soc} Calculation:} We exported the recorded voltage data for offline analysis. Although voltage was measured throughout the trip with a 1-second sampling rate, only the first and last data points were used to calculate the total \ac{soc} drop, representing the overall energy consumption for the trip. A nonlinear equation \eqref{eq: v2soc}, customised for lithium polymer battery characteristics \cite{Plett2004, RahimiEichi2013}, is then used to establish a precise correlation between voltage $V$ and \ac{soc}. The \ac{ebike} used in our study is equipped with a lithium polymer battery. We illustrate the characteristics of one battery cell, i.e., \textit{Tenergy 18650 battery cell}\footnote{\url{https://www.tenergy.com/30005_datasheet.pdf}}, in \autoref{fig: soc}, where the blue and yellow lines represent the actual and fitted curves, respectively. The \ac{soc} parameters in the yellow curve are obtained by least squares fitting with the \textit{Tenergy 18650 battery cell} data specification.

\begin{equation} \label{eq: v2soc}                 
    V = k_0 + k_1 \cdot SoC + \frac{k_2}{SoC} + k_3 \cdot \text{ln(}SoC\text{)} + k_4 \cdot \text{ln(}1-SoC\text{)}
\end{equation}

\noindent \textbf{Energy Efficiency Calculations:} Energy consumption is measured by analysing the decrease in \ac{soc} relative to battery capacity. Energy efficiency (Wh/km) is calculated by dividing the consumed energy by the travel distance collected from the odometer, providing insights into battery usage.

\noindent \textbf{Weather Data Incorporation:} Historical weather data (wind speed, direction, and descriptions) from Weather25\footnote{\url{https://www.weather25.com/europe/ireland/leinster/dublin}} is integrated to enrich the dataset.

\noindent \textbf{Synthetic Data Generation:} To overcome the limitations of our small real-world dataset, we employed synthetic data generation to augment our training set. Python's Synthetic Data Vault library\footnote{\url{https://docs.sdv.dev/sdv}}, which has been applied in many studies \cite{Kiran2023, Specht2023, Visani2022}, is used in our work to generate a 10,000-record supplement to the established real-world dataset. This library uses a variety of \ac{ml} algorithms, such as Gaussian Copula and CTGAN, to learn data patterns to generate synthetic data, enhancing the robustness and generalisation of data-driven models. Through rigorous validation to ensure consistency with the original dataset, we found that the quality of the generated data was 80.34\%. 

\subsubsection{\ac{escooter} Data}

The key steps for data processing are:

\noindent \textbf{Digit Extraction:} As the position of the digits representing \ac{soc} in the screen recording video is fixed, we selected this area as the \ac{roi}, as shown in the yellow box in \autoref{subfig: mi home interface}. When pixels in \ac{roi} change, we mark the previous and next frames that changed as keyframes and extract the digits by pytesseract\footnote{\url{https://github.com/madmaze/pytesseract}}, an optical character recognition tool for Python. We then conduct manual validation on the extracted data to guarantee its accuracy. The extracted digits are stored along with the corresponding video progress for data alignment.

\noindent \textbf{Data Alignment:} Data from two mobile applications is integrated using timestamps, a common attribute in \ac{gps} data and screen recordings. By adjusting the \ac{roi} and labelling keyframes with video progress, we can accurately correspond video to timestamps, ensuring precise \ac{soc} changes and effective data integration.

\noindent \textbf{Weather Data Incorporation \& Data Generation:} We employ the same methods described for \ac{ebike} data to enrich the dataset with weather information and generate a 10,000-record supplement to the established real-world dataset. The overall quality score of the generated data reaches 81.23\%.

\begin{figure*}[htbp]
    \vspace{-0.1in}
    \centering
    \subfigure[One sample \ac{escooter} trip]{
        \includegraphics[width=0.31\linewidth]{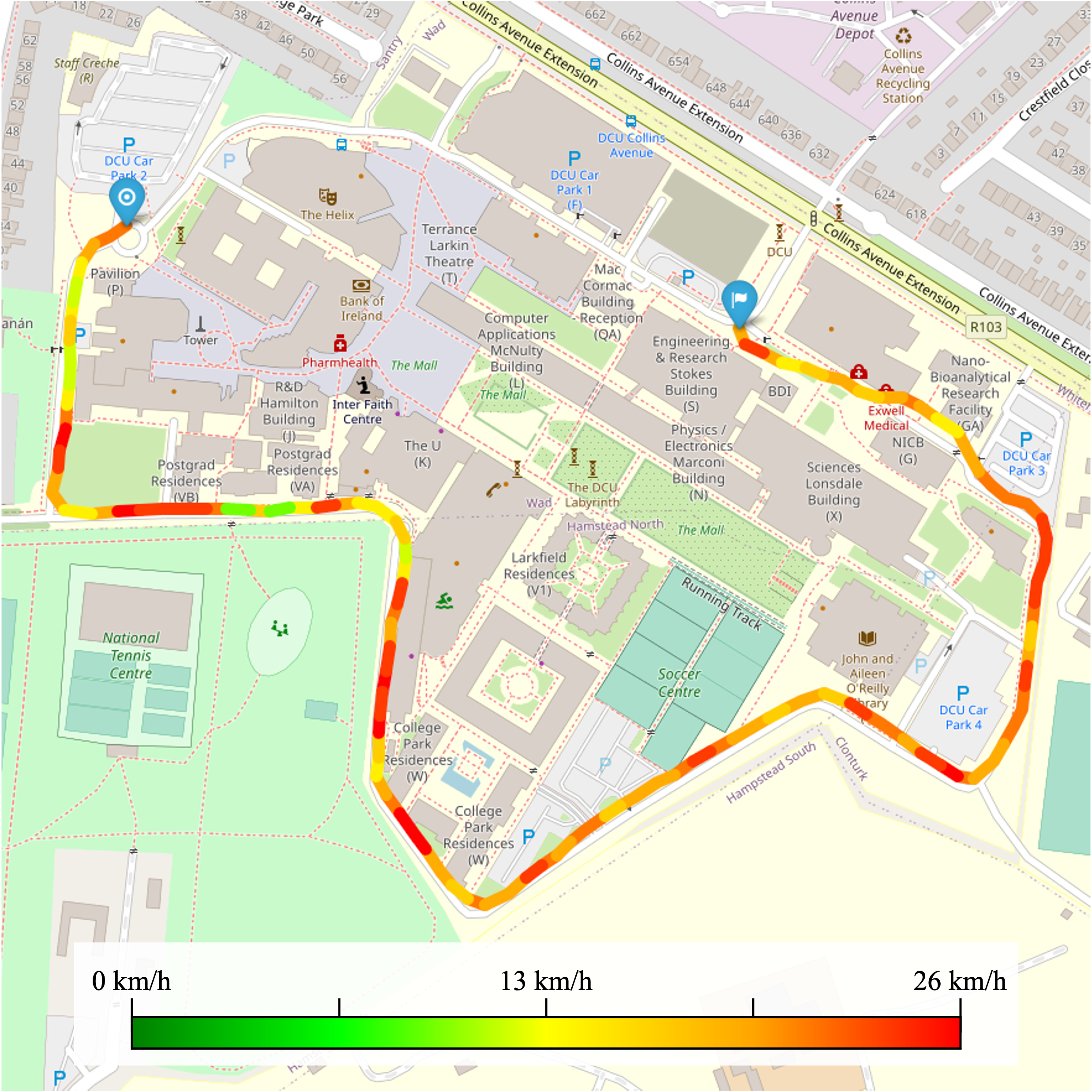}
        \label{subfig: scooter trip}
    }
    \subfigure[\ac{escooter} data analysis]{
        \includegraphics[width=0.62\textwidth]{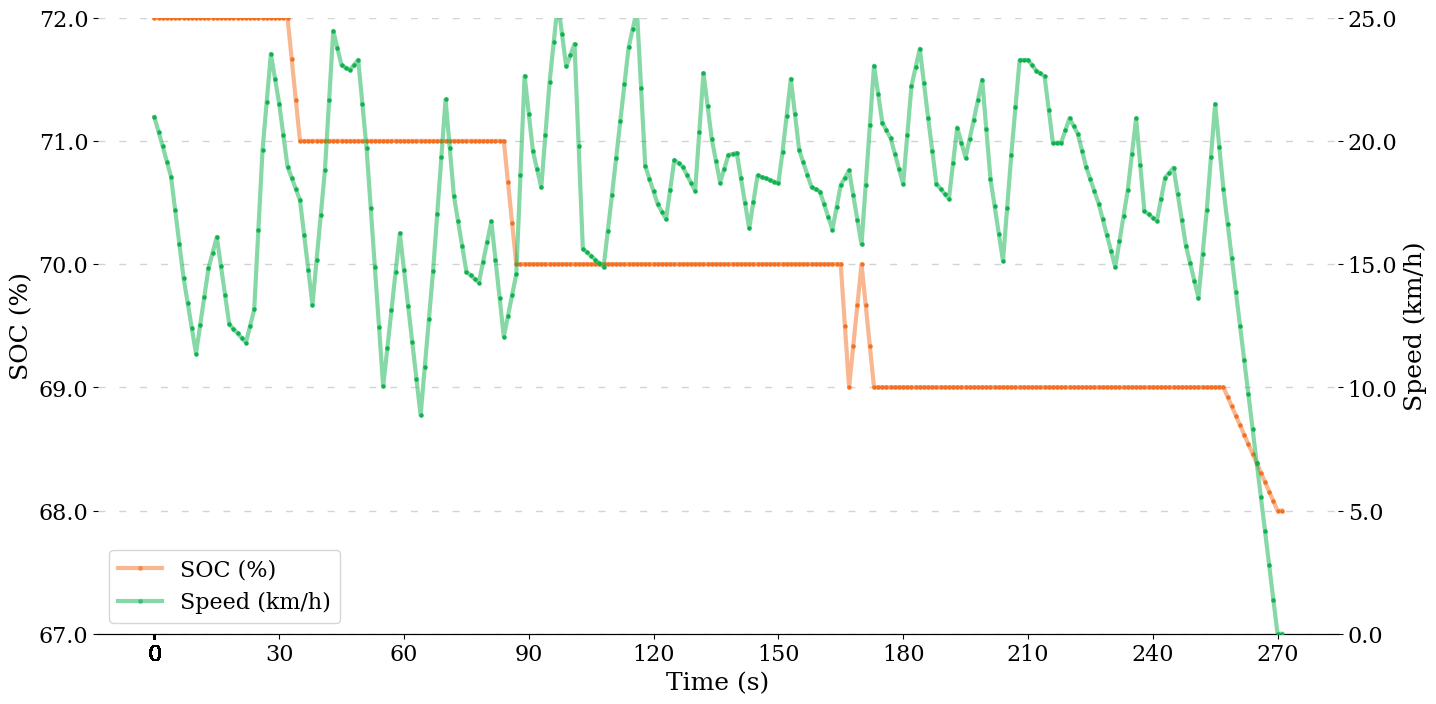}
        \label{subfig: scooter analysis}
    }
    \caption{Data analysis and visualisation of partial trajectory (starting from the 50th sample point) in \ac{escooter} Trip 27.}
    \vspace{-0.1in}
    \label{fig: scooter}
\end{figure*}

The dataset is primarily composed of two parts: specific driving data for each journey and a comprehensive summary table for all trips. Our study results in a dataset consisting of two parts: 36 \ac{ebike} trips and 30 \ac{escooter} trips. \autoref{table: ebike} and \autoref{table: escooter} provide an overview of some features from this collected data. A sample \ac{escooter} trip (Trip 27) is analysed and visualised in \autoref{fig: scooter}, in which \autoref{subfig: scooter trip} shows the speed changes along the trip trajectory, while \autoref{subfig: scooter analysis} presents the speed (green line) and \ac{soc} (orange line) trends over time. For access to the complete dataset, we suggest interested readers visit our GitHub repository\footnote{\url{https://github.com/SFIEssential/DualEMobilityData-datasets}}.

\begin{table*}[htbp]
    \caption{\ac{ebike} Dataset}
    \vspace{-0.1in}
    \label{table: ebike}
    \begin{tabularx}{\linewidth}{@{\extracolsep{\fill}}c c c c c c c c c c c}
        \toprule
        \textbf{timestamp} & \textbf{latitude} & \textbf{longitude} & \textbf{altitude} (m) & \textbf{speed} (km/h) & \textbf{wind speed} (km/h) & \textbf{wind direction} & \textbf{weather} & \textbf{temperature} (C) \\
        \midrule
        06/07/2023 09:04:23 & 53.3854 & -6.2564 & 56.8 & 18 & 16.9 & S & Cloudy & 16 \\
        06/07/2023 09:04:24 & 53.3854 & -6.2564 & 56.8 & 18 & 16.9 & S & Cloudy & 16 \\
        06/07/2023 09:04:25 & 53.3854 & -6.2565 & 56.8 & 18 & 16.9 & S & Cloudy & 16 \\
        06/07/2023 09:04:26 & 53.3854 & -6.2565 & 56.8 & 18 & 16.9 & S & Cloudy & 16 \\
        06/07/2023 09:04:27 & 53.3855 & -6.2566 & 56.8 & 18 & 16.9 & S & Cloudy & 16 \\
        \bottomrule
    \end{tabularx}
    \vspace{-0.1in}
\end{table*}

\begin{table*}[htbp]
    \caption{\ac{escooter} Dataset}
    \vspace{-0.1in}
    \label{table: escooter}
    \begin{tabularx}{\linewidth}{@{\extracolsep{\fill}}c c c c c c c c c}
        \toprule
        \textbf{timestamp} & \textbf{latitude} & \textbf{longitude} & \textbf{speed} (km/h) & \textbf{altitude} (m) & \textbf{\ac{soc}} (\%) & \textbf{wind speed} (km/h) & \textbf{wind direction} & \textbf{weather} \\
        \midrule
        12/10/2023 13:35:24 & 53.3916 & -6.2637 & 0.0000 & 57.2203 & 97 & 3 & E & Dry \\
        12/10/2023 13:35:31 & 53.3915 & -6.2633 & 2.1430 & 56.8687 & 97 & 3 & E & Dry \\
        12/10/2023 13:35:48 & 53.3915 & -6.2633 & 0.0000 & 56.8241 & 97 & 3 & E & Dry \\ 
        12/10/2023 13:36:06 & 53.3915 & -6.2634 & 8.1311 & 56.7919 & 97 & 3 & E & Dry \\
        12/10/2023 13:36:11 & 53.3915 & -6.2636 & 8.0638 & 55.9300 & 97 & 3 & E & Dry \\
        \bottomrule
    \end{tabularx}
    \vspace{-0.1in}
\end{table*}

\section{Models and Experiments} \label{ecm}
In this section, we introduced the experiments using mathematical and data-driven models to forecast energy consumption, discuss the results and analyse the implications for energy management in \ac{em} devices.

\subsection{Mathematical Model}

In our study, we adopted the mathematical energy consumption model initially proposed in \cite{Burani2022}, customising it to align with real-world application scenarios specific to our research. According to \cite{steyn2014comparison, upadhya2014characteristics, bertucci2013evaluation}, it is recognised that motor power is mainly affected by road slope, friction, and air resistance, and we can ascertain some physical parameters. We use $m$, $s$, and $v$ to represent the mass of the device and user, the road slope, and the driving speed respectively. The acceleration of gravity $g$ is approximately 9.81 $\mathrm{m/s^2}$. For asphalt surfaces, the rolling resistance coefficient $C_r$ is typically 0.001. The air density $\rho$ is given as 1.29 $\mathrm{kg/m^3}$. Additionally, the combined frontal area $A$ of the device and rider is 0.5 $\mathrm{m^2}$, and the drag coefficient $C_d$ is 0.7. To calculate the energy consumption per unit distance, we further refined the model and arrived at equation \eqref{eq: math definition}, which serves as the total energy required for \ac{escooter}, i.e., $P_{es} = P_{d}$. For \ac{ebike}, we introduce the power assistance level $P_{al}$, which ranges from 0 (no assistance) to 1 (full assistance), and formulate our prediction target, i.e., energy consumption demand per unit distance, as $P_{eb} = P_{d} \cdot P_{al}$.
        
\begin{equation} \label{eq: math definition}                 
    P_{d} = g \cdot m \cdot s + C_r \cdot m \cdot g + \frac{1}{2} C_d \cdot \rho \cdot A \cdot v^2
\end{equation}

\subsection{Data-Driven Models}

Nine data-driven algorithms, including \ac{lr}, \ac{svr}, \ac{dt}, \ac{rf}, \ac{gb}, \ac{knn}, \ac{xgb}, \ac{lgbm}, and \ac{mlp}, are chosen to develop predictive models for energy consumption in \ac{ebike}s and \ac{escooter}s. Each algorithm is selected based on its unique analytical strengths and attributes, which are essential for addressing the complexities and various factors influencing energy usage.

\subsection{Experiment Setup}

Our experimental datasets for both e-bikes and e-scooters contain the following feature categories:

\noindent \textbf{Rider Features:} height range (cm), weight range (kg).

\noindent \textbf{Trip Features:} \textit{1) \ac{ebike}:} distance (km), average speed (km/h), total ascent (m), average slope (\%), power assistance level; \textit{2) \ac{escooter}:} average speed (km/h), altitude difference (m), average slope (\%).

\noindent \textbf{Weather Features:} \textit{1) \ac{ebike}:} wind speed (km/h), wind direction, weather description, precipitation (mm), temperature (\textdegree C); \textit{2) \ac{escooter}:} wind speed (m/s), wind direction, weather description.

To process non-numeric features, we employed the following techniques:

\noindent \textbf{Label Encoding:} distinct numerical labels were assigned to textual weather descriptions, such as ``Dry" and ``Wet".

\noindent \textbf{Vectrisation:} wind direction was converted into two numerical columns, \textit{WE} and \textit{NS}, using the Cartesian coordinate system, with the positive directions of the $x$ and $y$ axes representing east and north respectively.

We initially applied the \ac{lr}, \ac{svr}, \ac{dt}, \ac{rf}, \ac{gb}, \ac{knn}, \ac{xgb} and \ac{lgbm} to implement the prediction on two datasets, which were both divided into training, testing, and validation sets with an 8:1:1 ratio. Then, the \ac{mlp} model was designed and optimised by the \textit{Adam} optimiser with a default learning rate of $1e^{-3}$. The \ac{xgb} and \ac{lgbm} regressors were implemented using their respective Python libraries, while other models were built using \textit{Tensorflow Keras} and \textit{Sklearn} libraries. 

We conducted experiments both with and without rider features (height and weight ranges) to assess their impact on prediction results. While these features cannot be removed from the mathematical consumption model due to their physical significance, their influence on \ac{ml} model performance was analysed. In this study, Watt-hours per kilometre (Wh/km) is the primary metric for assessing energy consumption efficiency in electric vehicles. Lower Wh/km values indicate better efficiency. Model performance was assessed based on their \ac{mae} in predicting energy consumption, and lower \ac{mae} values signify higher accuracy.

\section{Results and Discussion} \label{rd}

\begin{table*}[ht]
    \caption{Performance (MAE) of distinct approaches (Wh/km) before and after feature removal.}
    \vspace{-0.1in}
    \label{table: performance}
    \begin{tabularx}{\linewidth}{@{\extracolsep{\fill}}c c c c c}
        \toprule
        \multirow{2}{*}{\textbf{Approach}} & \multicolumn{2}{c}{\textbf{\ac{ebike}}} & \multicolumn{2}{c}{\textbf{\ac{escooter}}} \\
        \cmidrule(lr){2-3} \cmidrule(lr){4-5}
        & with user features & without user features & with user features & without user features \\
        \midrule
        Mathematical model & 29.39 & -- & 27.87 & -- \\
        \ac{lr} & 4.68 & 4.84 & 4.85 & 6.44\\
        \ac{svr} & 4.66 & 4.80 & 4.87 & 6.40 \\
        \ac{dt} & 6.03 & 6.49 & 6.85 & 9.09\\
        \ac{rf} & 4.63 & 4.74 & 4.98 & 6.57 \\
        \ac{gb} & 4.59 & 4.73 & 4.83 & 6.44\\
        \ac{knn} & 5.47 & 5.42 & 5.40 & 6.95 \\
        \ac{xgb} & 4.69 & 4.86 & 5.13 & 6.84 \\
        \ac{lgbm} & 4.57 & 4.70 & 4.82 & 6.55\\
        \ac{mlp} & \textbf{4.47} & \textbf{4.62} & \textbf{4.77} & \textbf{6.32} \\
        \bottomrule
    \end{tabularx}
    \vspace{-0.1in}
\end{table*} 

The performance of various predictive models in estimating the energy consumption of \ac{ebike} and \ac{escooter} is compared in \autoref{table: performance}, measured in Watt-hours per kilometre. In each experiment, the \ac{mlp} model consistently exhibits superior performance, as evidenced by its \ac{mae} values of 4.47, 4.62, 4.77, and 6.32, surpassing other models. This result suggests that while advanced models like \ac{mlp} can achieve remarkable accuracy, the choice of model should be tailored to the specific characteristics of the data. Following the removal of users' height and weight features, it shows  \autoref{table: performance} that the \ac{mae} values are consistently higher than those before removal for most models (it slightly dropped for \ac{knn}). This increase in \ac{mae} indicates a decline in prediction accuracy, underscoring the importance of user-specific features in the model's performance. 

\section{Limitations} \label{dl}
This section delves into the discussions surrounding our study's findings and critically evaluates the inherent limitations therein. We classify our limitations from the views of Electric Micromobility devices, data sources, contributors to data acquisition, and data collection process.

In our study, we did not compare devices across different models and vendors, hence our work should be regarded as an initial exploration in this domain. Additionally, our devices are relatively new, which may omit considerations for wear-and-tear or ageing effects, commonly referred to as 'friction' in the longevity of device performance.

Further, our verification of the collected data was limited; for instance, we did not use multiple \ac{gps} sensors, speed sensors, or altitude measuring instruments to cross-validate the location, altitude, and speed data. Moreover, we relied on external web sources for climate details, which provided data with coarse granularity (hourly), lacking finer temporal resolution that might affect the accuracy of our analysis. Also, our research was primarily focused on evaluating models utilising average speed as a basis for trip planning. However, we did not investigate detailed aspects of modelling beyond this parameter. This will be part of our future work. 

Finally, the data for our study was gathered through voluntary participation, which constrained the size of our dataset. To address this, we employed the Synthetic Data Vault method to augment our data pool. However, this method only simulates the distribution of the original data, which may not accurately reflect the diversity of potential user profiles. Additionally, our approach was not personalised to individual users, mainly due to compliance with the \ac{gdpr}, limiting our capability to carry out customised energy prediction models and optimisation based on data collected from any specific user for travel behavior analysis. For instance, the height and weight data were collected in ranges, which could affect the precision of personalised modelling. Additionally, our data collection was geographically restricted to Dublin, Ireland, and primarily conducted on university campuses due to health and safety considerations. This limited the diversity of environmental conditions in our data, potentially affecting the applicability of our findings across different terrains and temporal variations. 

\section{Conclusion and Future Work} \label{cf}

 Our research represents a pivotal step in bridging the data shortage in understanding the energy consumption of \ac{em}s. We introduce two meticulously collected open-source benchmark datasets for \ac{ebike} and \ac{escooter}, aiming to capture and analyse the usage patterns of these vehicles. This comprehensive dataset we proposed contributes to the current gap in relevant datasets with detailed trip information, especially energy consumption. Through in-depth experiments and analysis, we found that our \ac{mlp} model improved prediction accuracy for \ac{ebike}s and \ac{escooter}s. Also, the user-specific features are important for the improvement of models' performance. As part of our future work, we aim to explore the limitations mentioned to enrich our research and practical application. We will use \ac{gdpr}-compliant historical data to improve predictions and employ federated learning, as introduced in \cite{9625535, FedBEVJ}, to further enhance accuracy in model performance in a collaborative learning setup. We also plan to diversify our dataset with various terrains for robust, precise energy consumption forecasts and incorporate detailed trajectory data for real-time monitoring and in-depth data analysis.

\section*{Acknowledgment}

This work has emanated from research supported in part by Science Foundation Ireland under Grant Number \textit{21/FFP-P/10266} and \textit{SFI/12/RC/2289\_P2} (Insight the SFI Research Centre for Data Analytics, at Dublin City University). Yue Ding was also supported by the SFI Centre for Research Training in Machine Learning (ML-Labs) at Dublin City University. We extend special thanks to our Chief Technical Officer \textit{Paul Wogan} at the School of Electronic Engineering, Dublin City University, for his invaluable support in designing, equipping and integrating the voltage logger into our \ac{ebike}.

The studies involving human participants were reviewed and approved by Data Protection Office, Dublin City University. Written informed consent for participation was required for this study in accordance with the national legislation and the institutional requirements. The project has received ethical approval from Dublin City University Research Ethics Committee with reference number \textit{DCUREC/2023/025}.

\bibliographystyle{IEEEtran}
\bibliography{reference}

\end{document}